\documentclass[runningheads]{llncs}
\usepackage[T1]{fontenc}
\usepackage{graphicx}
\usepackage{booktabs}
\usepackage{tabularx}
\usepackage[whole]{bxcjkjatype}
\usepackage{tcolorbox}
\usepackage{color}

\begin{document}

\title{Synth-JDoc: Synthesizing a Japanese Document Image Dataset for OCR with Diverse Layouts and Embedded Images}

\titlerunning{Synth-JDoc}
\author{Keito Sasagawa\inst{1,3} \and
Shuhei Kurita\inst{2,3} \and
Daisuke Kawahara\inst{1,3}}

\authorrunning{K. Sasagawa et al.}

\institute{Waseda University, Tokyo, Japan \\
\email{\{kate@fuji.,dkw@\}waseda.jp} \and
NII, Tokyo, Japan \\
\email{skurita@nii.ac.jp} \and
NII LLMC, Tokyo, Japan\\
}
\maketitle
\begin{abstract}

The ability of Large Vision Language Models (LVLMs) to read text within document images is crucial, as it enables various applications such as Document Visual Question Answering.
To enhance the text-reading capabilities of LVLMs, high-quality OCR datasets are essential.
This need is particularly critical for Japanese documents, which often feature vertically written text alongside horizontally written text.
Current LVLMs demonstrate considerably lower performance on vertically written Japanese text than on horizontally written text, necessitating specialized OCR datasets to bridge this gap.
However, manually constructing OCR datasets is expensive and difficult to scale.
Alternatively, constructing datasets by extracting text from existing document images using OCR models introduces challenges, such as text recognition errors and the prerequisite of sourcing document images.

To address these issues, we construct an OCR dataset by synthesizing document images directly from text.
Leveraging HTML and CSS, we generate multi-column documents that incorporate both vertical and horizontal writing styles.
Furthermore, to ensure the visual realism of the documents, we embed images generated by text-to-image models within the layout.
Additionally, to foster model robustness, we apply noise and degradation filters to the synthesized document images.
In our experiments, we compared the performance of models fine-tuned on our synthetic dataset against baselines fine-tuned on synthetic datasets from prior work and those generated by a high-performance text-to-image model.
Evaluation results demonstrate that our synthetic dataset is the most effective approach for improving LVLM performance on reading vertically written Japanese text.
Our dataset and code are publicly available (https://github.com/llm-jp/synth-jdoc).

\end{abstract}

\section{Introduction}

\begin{figure}[h]
  \begin{minipage}[c]{0.21\columnwidth}
    \centering
    \includegraphics[keepaspectratio, width=\columnwidth]{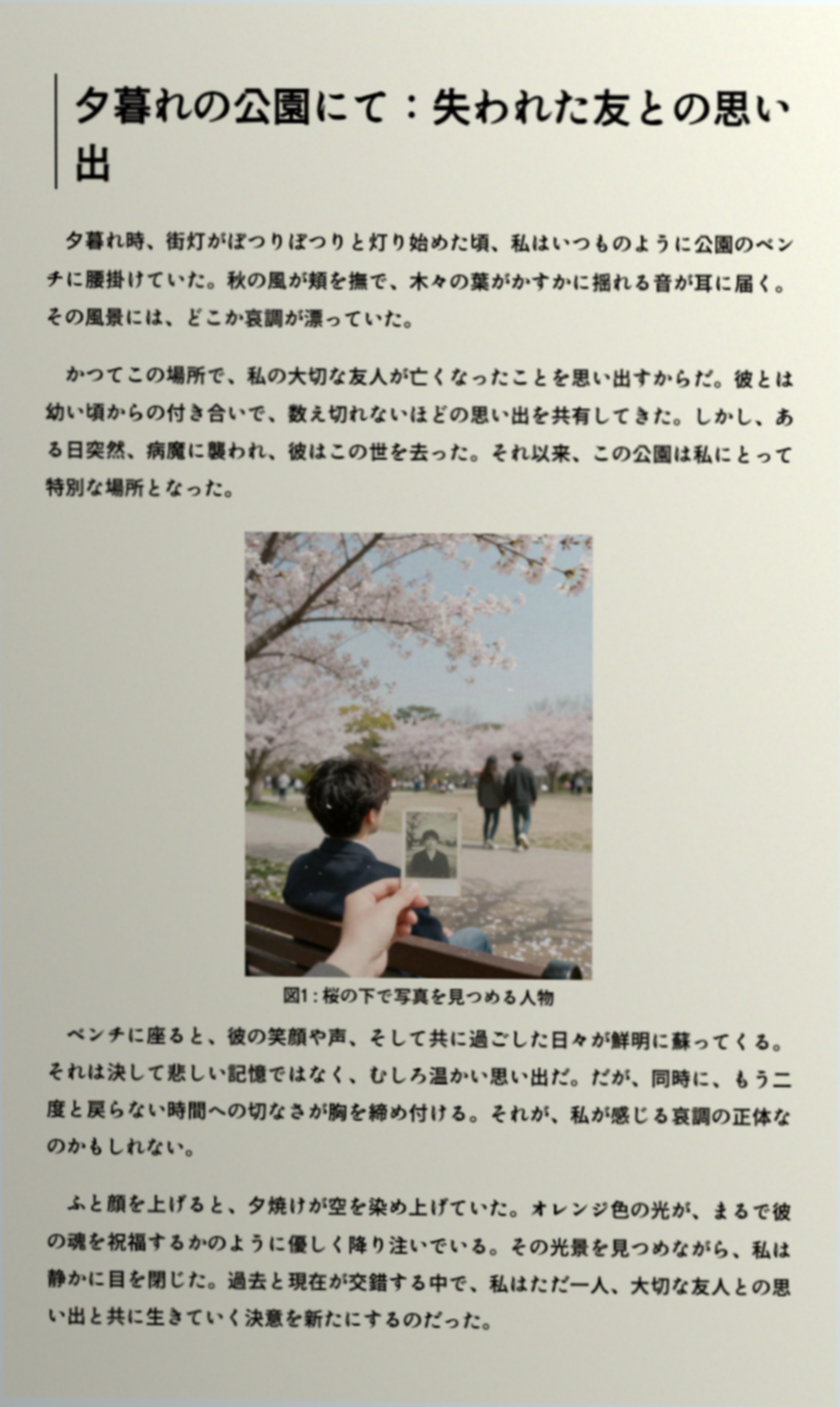}
  \end{minipage}
  \begin{minipage}[c]{0.38\columnwidth}
    \centering
    \includegraphics[keepaspectratio, width=\columnwidth]{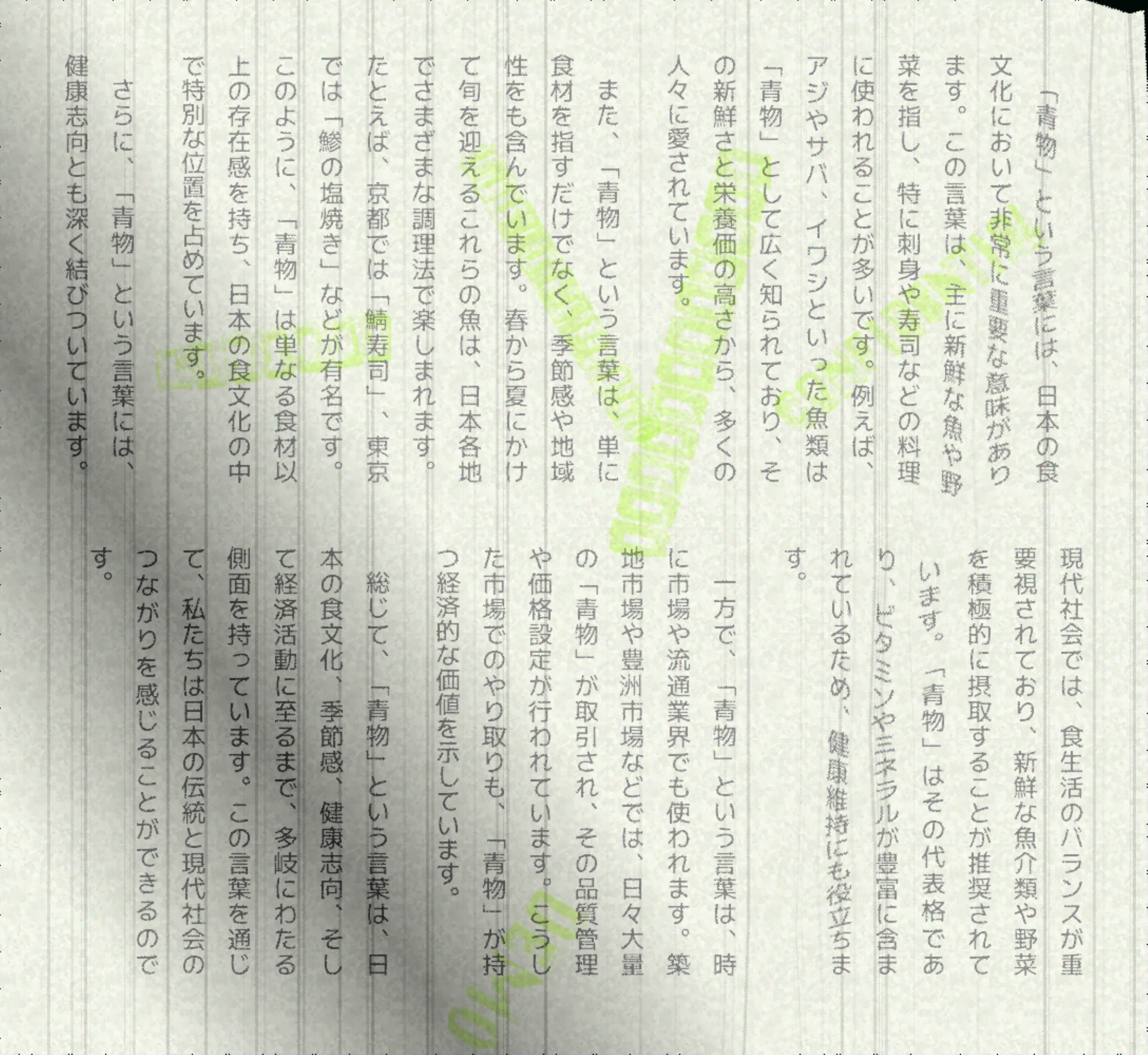}
  \end{minipage}
  \begin{minipage}[c]{0.38\columnwidth}
    \centering
    \includegraphics[keepaspectratio, width=\columnwidth]{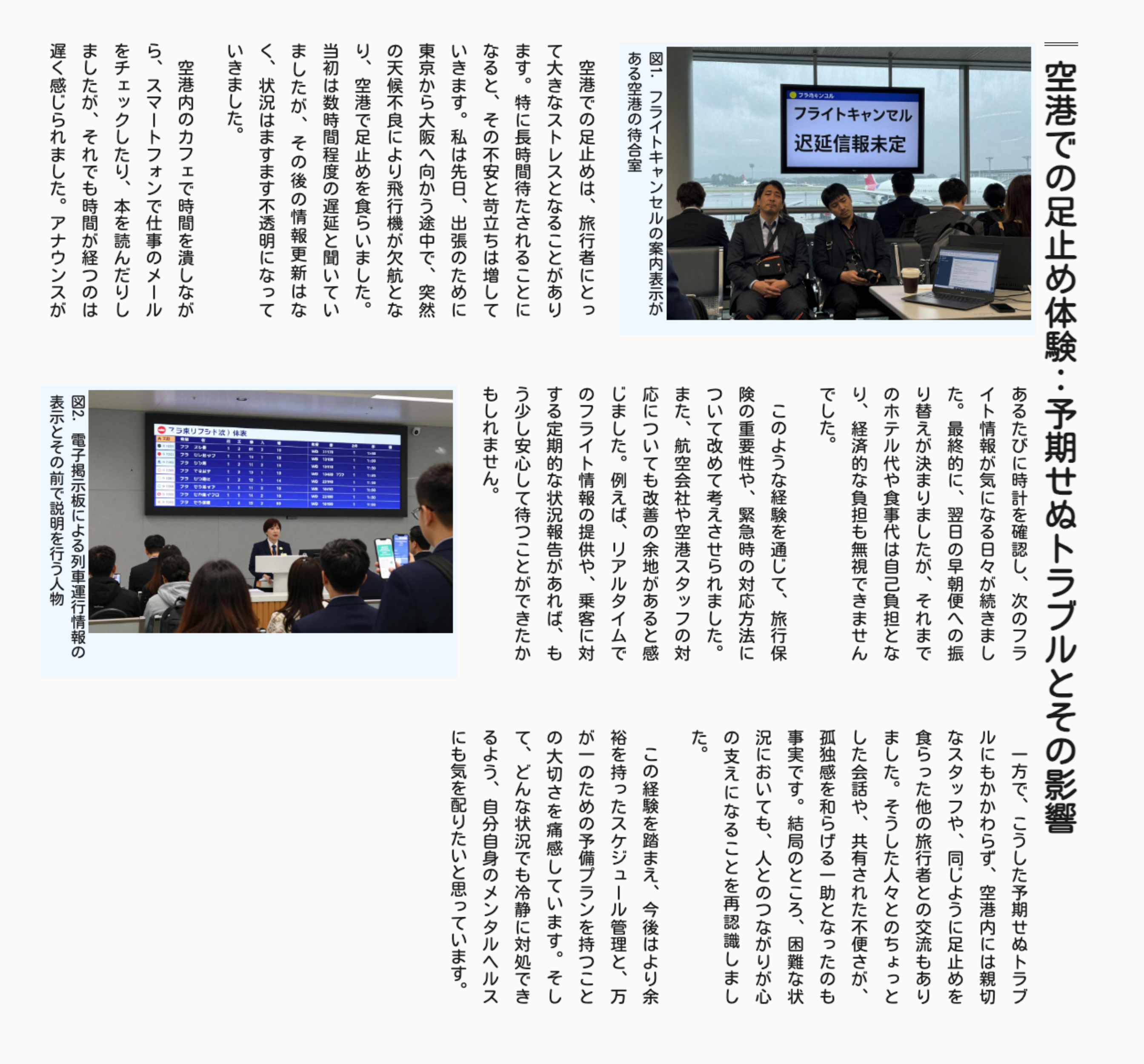}
  \end{minipage}\\
  \caption{
  Examples of document images from the \textbf{Synth-JDoc} dataset.
  The documents feature horizontally or vertically written text in 1- to 4-column layouts, with occasional image insertions.
  Noise is applied to a subset of the images to enhance model robustness.
  (Left: Horizontally written, single-column layout with an inserted image.
  Center: Vertically written, two-column layout without images.
  Right: Vertically written, three-column layout with inserted images.)
  }
  \label{fig:example_synth_jdoc}
\end{figure}

Large Vision Language Models (LVLMs), which can process both images and text to understand visual content and perform visual question answering, have rapidly advanced in recent years \cite{openai2025gpt5,google_gemini3_2026,bai2025qwen3vltechnicalreport,wang2025internvl35advancingopensourcemultimodal,gemmateam2025gemma3technicalreport}.
One of the critical downstream tasks for LVLMs is Document Visual Question Answering \cite{Mathew_2021_WACV,Tanaka_Nishida_Yoshida_2021,Mathew_2022_WACV,Tanaka_Nishida_Nishida_Hasegawa_Saito_Saito_2023,onami-etal-2024-jdocqa}, where the model answers questions based on document images.
Performing this task requires comprehending the document's content, which inherently necessitates the ability to read the text within the image.
To enhance this capability, OCR datasets for training LVLMs are essential.
In particular, alongside horizontally written text, there are also Japanese documents written vertically.
In vertically written Japanese documents, characters are read from top to bottom, and lines are read from right to left.
Additionally, in multi-column layouts, the reading order progresses sequentially from the uppermost column to the lowest.
Because few OCR datasets focus on vertically written Japanese, models currently exhibit relatively low proficiency in reading vertically written text \cite{sasagawa2025evaluatingmultimodallargelanguage}.
Therefore, constructing a Japanese document image OCR dataset that includes vertically written text is crucial.

Common approaches for constructing OCR datasets include manual annotation; however, this method incurs prohibitive costs and is difficult to scale.
Another approach involves extracting text from document images using existing OCR models, but this method suffers from OCR errors and the prerequisite burden of collecting document images.
Synthesizing document images from given text can address these issues, but previous document image synthesis methods fail to support the complex layouts of Japanese documents, such as those containing vertically written text and inserted images.

To tackle these limitations, we propose a novel Japanese document image synthesis method and construct the \textbf{Synth-JDoc} dataset.
First, we prepare the text to be rendered within the document images and segment it into paragraphs.
For selected paragraphs, we generate corresponding images using a text-to-image model, and subsequently generate captions for these synthesized images using an LVLM.
Additionally, an appropriate title is generated from the prepared text using a Large Language Model (LLM).
Using these elements, we synthesize document images featuring vertically or horizontally written, multi-column layouts via HTML and CSS.
Furthermore, we apply noise to a portion of the images to enhance model robustness.

In our experiments, we validated the effectiveness of the constructed Synth-JDoc using a real-world Japanese document OCR dataset that features vertically written text.
We compared the performance of models fine-tuned on Synth-JDoc against several baseline settings.
The baselines included the original models without fine-tuning, models fine-tuned on datasets from prior research, and models fine-tuned on synthetic document image datasets generated by a high-performance text-to-image model.
The evaluation results demonstrate that training on Synth-JDoc most effectively improves OCR capabilities on Japanese document images.

The dataset and the code used for our experiments are publicly available\footnote{code: https://github.com/llm-jp/synth-jdoc\\ dataset: https://huggingface.co/datasets/llm-jp/Synth-JDoc}.

\section{Related Work}
\subsection{Synthetic Document Dataset}

Methods that synthesize document images from provided text offer the advantages of eliminating the need for document image collection and avoiding OCR errors.
Genalog \cite{gupte2021genalog} is a library that synthesizes realistic document images from given text using HTML and CSS and generates degraded versions of these images.
Genalog can generate horizontally written documents using three templates: multi-column layouts resembling academic papers, letter-style layouts, and simple text-block layouts.
It then applies realistic degradation, such as Gaussian blur, to the generated images.
DocCreator \cite{jimaging3040062} utilizes a small number of real document images to extract three key elements: font, background, and layout. It then uses these elements to synthesize document images containing ground truth text.
Subsequently, it performs data augmentation by applying realistic degradation to the synthesized images.
SynthDoG \cite{10.1007/978-3-031-19815-1_29} generates images by rendering words or phrases onto paper textures and compositing them with background images.
While this method can generate Japanese document images, the resulting layouts lack realism.
SynthDoc \cite{10.1145/3688866.3689125} synthesizes document images from prepared text and images through two distinct processes: Layout Design and Content Rendering.
The Layout Design process determines the placement of elements within the document, while the Content Rendering process renders elements such as images and text. JSSODa \cite{sasagawa2025evaluatingmultimodallargelanguage} is a synthetic Japanese document image dataset that includes both vertically and horizontally written text with one to four columns.
This dataset is constructed by rendering black text on a white background, without any inserted images or figures.

These existing methods are incapable of synthesizing document images with complex layouts that feature vertically written text, multiple columns, and inserted images.

\subsection{Japanese Document Dataset}

JDocQA \cite{onami-etal-2024-jdocqa} is a QA dataset designed for Japanese document images, constructed by manually annotating QA pairs on real-world PDF pages.
CC-OCR \cite{Yang_2025_ICCV}, a test dataset for evaluating the OCR capabilities of LVLMs, contains approximately ten camera-captured Japanese document images.
Additionally, VJRODa \cite{sasagawa2025evaluatingmultimodallargelanguage} is an OCR test dataset specifically for vertically written Japanese document images, constructed from real-world PDF pages.
This dataset comprises 100 pairs of document images and their corresponding texts.

Regarding OCR training datasets for Japanese documents, synthdog-ja \cite{10.1007/978-3-031-19815-1_29} has been developed.
This dataset consists of Japanese document images generated by SynthDoG; however, it suffers from unnatural layouts.
Similarly, while JSSODa \cite{sasagawa2025evaluatingmultimodallargelanguage} is a synthetic Japanese document image dataset that includes both vertically and horizontally written text with one to four columns, it lacks visual diversity because no images are inserted within the documents.
Consequently, there is a significant shortage of comprehensive OCR training datasets for Japanese documents, a gap that we address in this study.

\section{Synth-JDoc: Synthetic Japanese Document Dataset}

\begin{figure}[t]
    \begin{center}
    \includegraphics[width=\columnwidth]{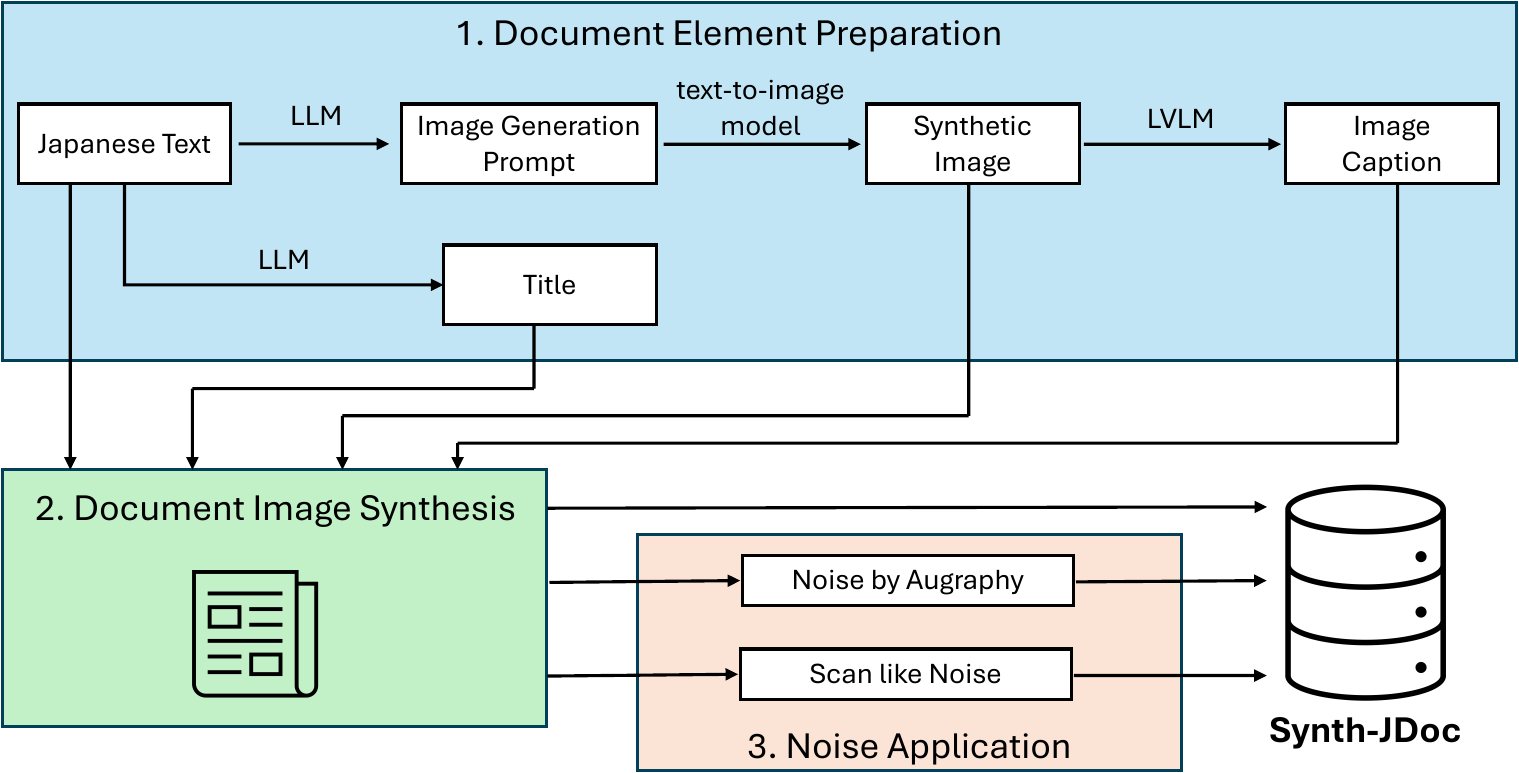}
    \caption{
    Overview of the \textbf{Synth-JDoc} Construction Pipeline
    }
    \label{fig:overview}
    \end{center}
\end{figure}

We synthesize Japanese document images with diverse layouts using HTML and CSS.
Because our method synthesizes document images by generating document elements, such as images and titles, exclusively from text, it offers the distinct advantage of eliminating both the need for pre-existing document images and the occurrence of OCR errors.
Furthermore, by controlling the layout via HTML and CSS, we can synthesize document images featuring diverse layouts, including vertically or horizontally written multi-column structures with inserted images.
Fig. \ref{fig:overview} illustrates an overview of the proposed image synthesis method.
The pipeline comprises three distinct stages: document element preparation, document image synthesis, and noise application.

During the document element preparation stage, we first prepare the Japanese text to be rendered within the document images and segment it into paragraphs.
For selected paragraphs, corresponding images are generated using a text-to-image model.
Subsequently, an LVLM is employed to generate captions for these synthesized images.
Additionally, a document title is generated based on the entire text.
In the document image synthesis stage, these prepared elements are utilized to construct web pages via HTML and CSS, which are then rendered into images.
Finally, in the noise application stage, noise is applied to the images to enhance robustness.

\subsection{Document Element Preparation}
\label{ssec:element_preparation}
\subsubsection{Preparing text and title}

We utilize text from the JSSODa dataset~\cite{sasagawa2025evaluatingmultimodallargelanguage} as the content to be rendered within the document images.
The texts in this dataset were generated by inputting nouns extracted from a Japanese dictionary (the JUMAN dictionary\footnote{https://github.com/ku-nlp/JumanDIC}) into the llm-jp-3.1-13b-instruct4 model\footnote{https://huggingface.co/llm-jp/llm-jp-3.1-13b-instruct4} \cite{llmjp2024llmjpcrossorganizationalprojectresearch}, instructing it to generate sentences related to each noun.
For each text, we generated an appropriate title using llm-jp-3.1-13b-instruct4.

\subsubsection{Image Generation for Embedding in Documents}

We segmented each text into paragraphs by splitting it at ``\textbackslash n\textbackslash n''.
The number of images to insert into a single document image was randomly determined, ranging from a minimum of zero to a maximum of half the total number of paragraphs.
Next, we randomly inserted image placeholders either before or after the paragraphs and assigned each placeholder to a corresponding paragraph.
We formulated this assignment to minimize the sum of squared distances between the paired image placeholders and paragraphs.
To determine these optimal pairs, we constructed a cost matrix by calculating the squared distance for each potential pair and solved the linear assignment problem.

Subsequently, based on the paragraph text assigned to each image placeholder, we generated an image generation prompt relevant to its content using Qwen3-30B-A3B-Instruct-2507\footnote{https://huggingface.co/Qwen/Qwen3-30B-A3B-Instruct-2507} \cite{yang2025qwen3technicalreport}.
We then generated the corresponding images using Z-Image-Turbo\footnote{https://huggingface.co/Tongyi-MAI/Z-Image-Turbo} \cite{imageteam2025zimageefficientimagegeneration}, a text-to-image model, with these prompts.
Furthermore, we employed Qwen3-VL-30B-A3B-Instruct\footnote{https://huggingface.co/Qwen/Qwen3-VL-30B-A3B-Instruct} \cite{bai2025qwen3vltechnicalreport} to generate captions for the synthesized images.
To ensure layout diversity, we generated concise captions and lengthy, multi-sentence captions at a ratio of 9:1.

\subsection{Document Image Synthesis}

We synthesized the document images using the prepared document elements.
For each synthesized image, the orientation (horizontal or vertical) and the number of columns (ranging from 1 to 4) were set identical to those in JSSODa.
The generated images are uniformly distributed across the eight possible combinations of orientation and column counts.
The document title is displayed with a 25\% probability.
When inserting generated images into layouts with two or more columns, we created configurations where the image is either confined within a single column or spans across all columns.
Furthermore, various caption styles for the inserted images (e.g., ``Figure N:'', ``Fig. N:'') were designed and applied randomly.
We randomly selected the font from 49 Japanese options available on Google Fonts\footnote{https://fonts.google.com/}.
Incorporating these variations, we constructed HTML/CSS templates and rendered the images using a web browser.
Examples of the synthesized images are shown in Fig. \ref{fig:example_synth_jdoc}.

\begin{figure}[h]
  \begin{minipage}[c]{0.1\columnwidth}
    Clean
  \end{minipage}
  \hfill
  \begin{minipage}[c]{0.33\columnwidth}
    \centering
    \includegraphics[keepaspectratio, width=\columnwidth]{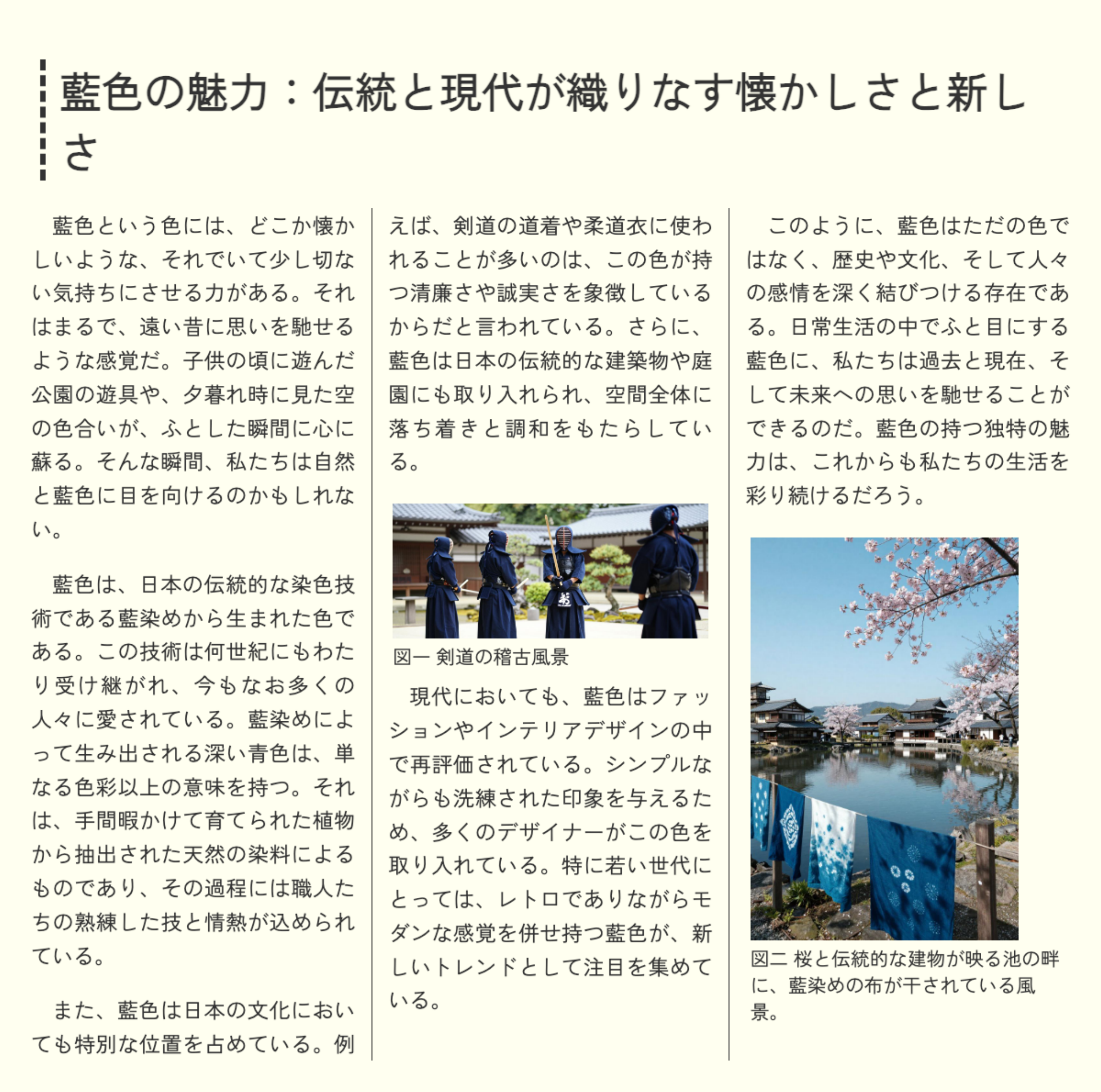}
  \end{minipage}
  \begin{minipage}[c]{0.5\columnwidth}
    \centering
    \includegraphics[keepaspectratio, width=\columnwidth]{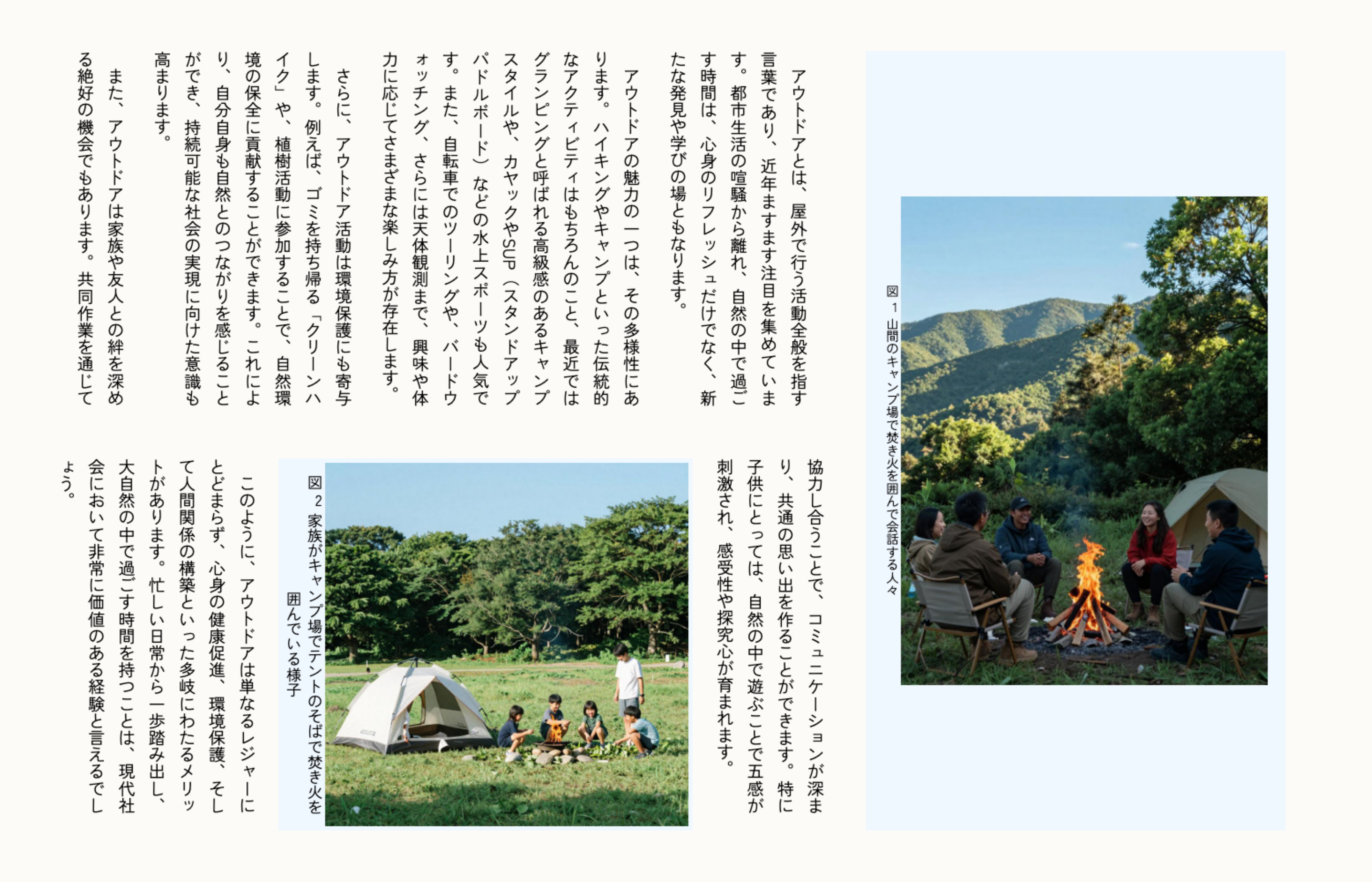}
  \end{minipage}\\

  \begin{minipage}[c]{0.1\columnwidth}
    Scan
  \end{minipage}
  \hfill
  \begin{minipage}[c]{0.33\columnwidth}
    \centering
    \includegraphics[keepaspectratio, width=\columnwidth]{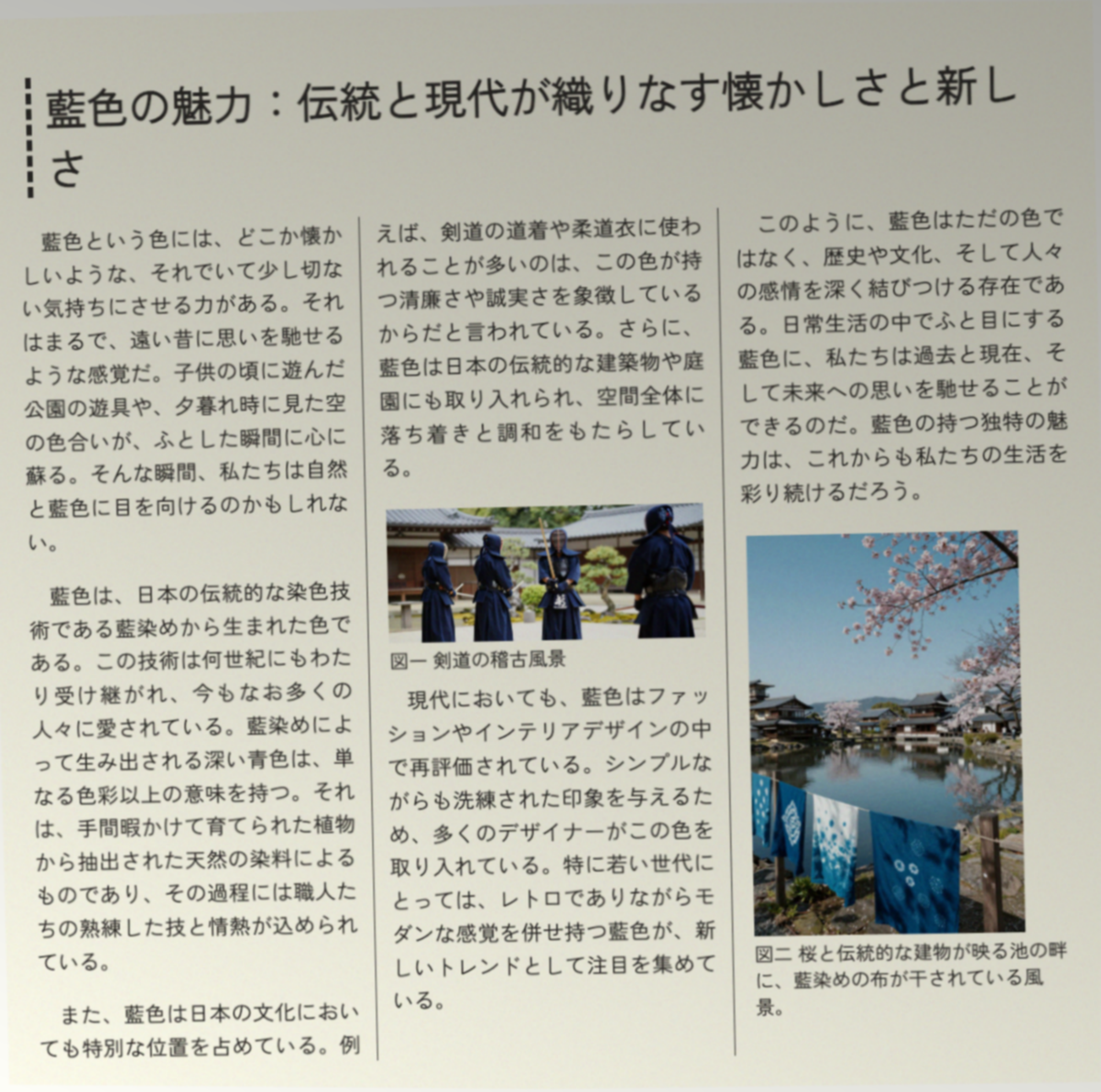}
  \end{minipage}
  \begin{minipage}[c]{0.5\columnwidth}
    \centering
    \includegraphics[keepaspectratio, width=\columnwidth]{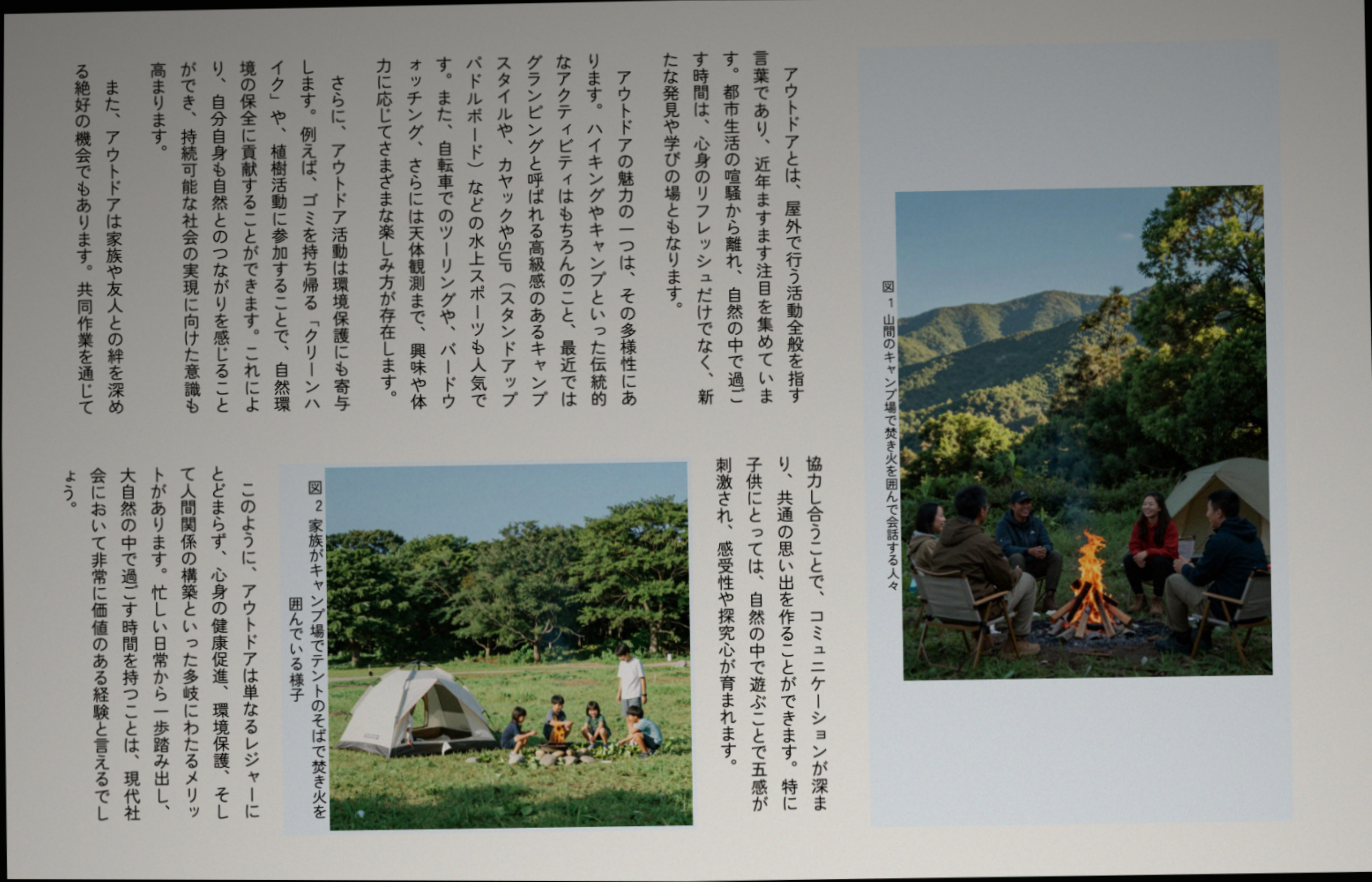}
  \end{minipage}\\

  \begin{minipage}[c]{0.1\columnwidth}
    Augraphy
  \end{minipage}
  \hfill
  \begin{minipage}[c]{0.33\columnwidth}
    \centering
    \includegraphics[keepaspectratio, width=\columnwidth]{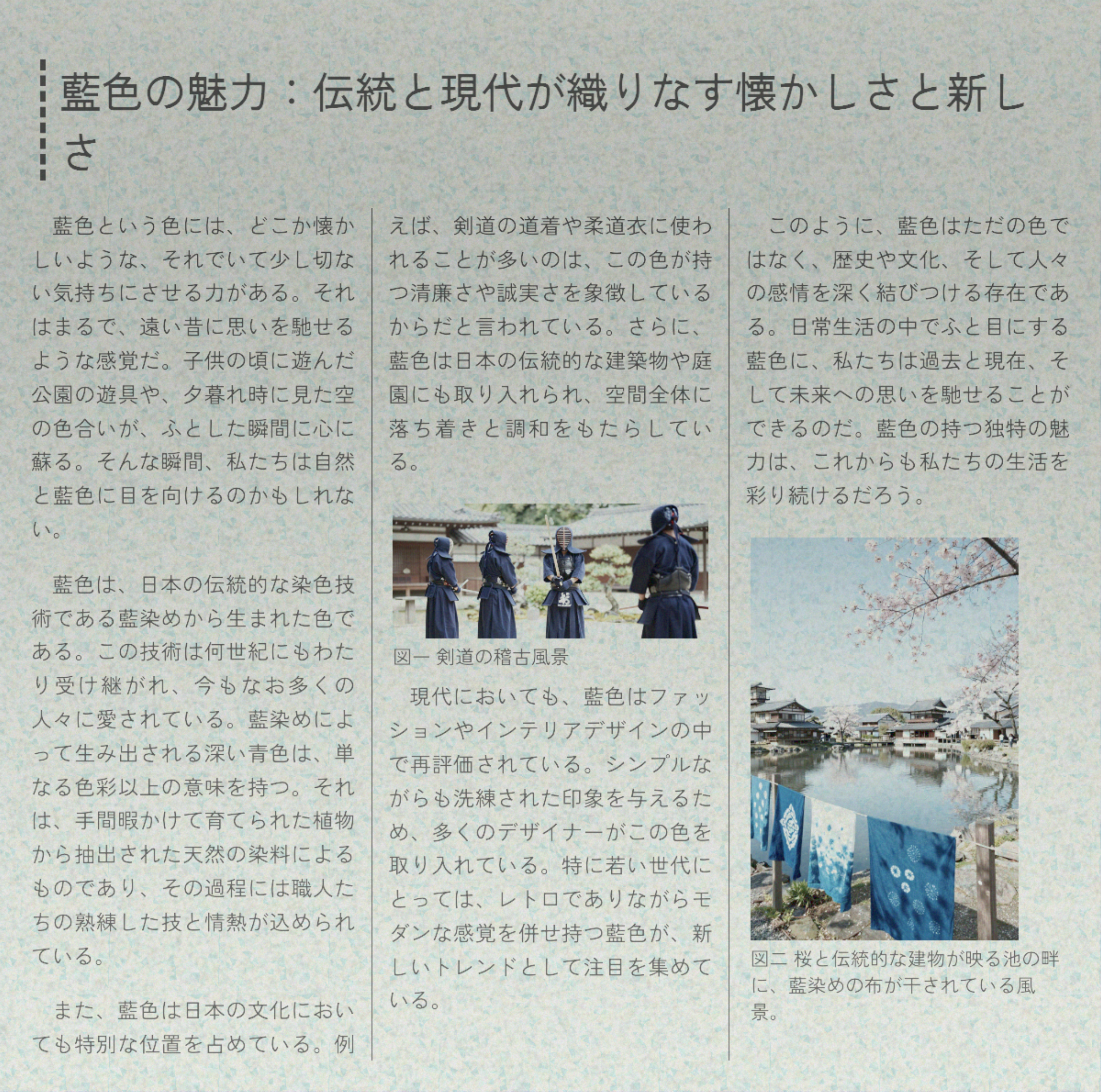}
  \end{minipage}
  \begin{minipage}[c]{0.5\columnwidth}
    \centering
    \includegraphics[keepaspectratio, width=\columnwidth]{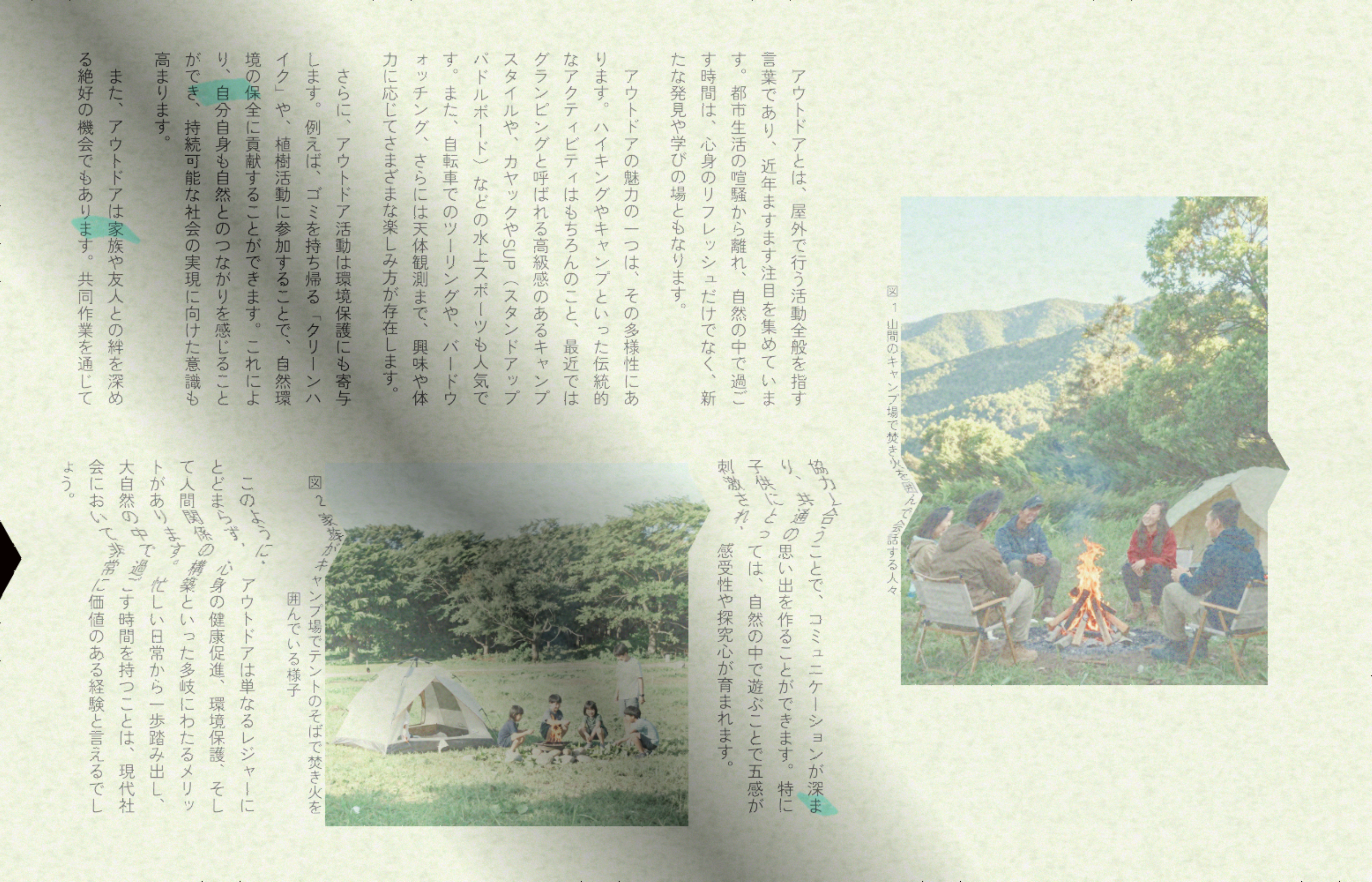}
  \end{minipage}\\
  \caption{
  Example images illustrating each noise type.
  Rows indicate noise types: ``Clean'' (clean document), ``Scan'' (scan-like noise), and ``Augraphy'' (Augraphy library).
  }
  \label{fig:example_noise}
\end{figure}

\subsection{Noise Application}

We apply noise to the synthesized document images using two distinct methods.
The first method transforms the images to resemble scanned documents.
The second method introduces artificial degradation using the Augraphy library~\cite{augraphy_paper,augraphy_library} to enhance robustness.

\subsubsection{Noise like Scanned Images}

We transform the synthesized document images to resemble scanned documents.
First, to simulate the texture of real paper, we apply Gaussian noise across the entire image.
To replicate the skew and distortion that occur during scanning, we apply rotation and perspective transformations to the images.
Additionally, to emulate shadows introduced during the scanning process, we randomly apply vertical and horizontal linear shadows, as well as vignette effects.
Finally, to represent text blurring, we apply a global blur effect with a certain probability.
Examples of the resulting images are shown in Fig. \ref{fig:example_noise}.

\subsubsection{Noise by Augraphy}

To enhance the robustness of the model's OCR capabilities, we utilize the Augraphy library to generate visually challenging document images.
The Augraphy pipeline is divided into three distinct stages: the Ink Phase, the Paper Phase, and the Post Phase.

The Ink Phase targets the degradation of characters within the image; in this study, we apply the InkBleed and InkMottling effects, each with a certain probability.
The subsequent Paper Phase focuses on degrading the background paper.
Here, we apply processes that add color and textural patterns to the paper.
Finally, Post Phase operations are applied to the output images from the preceding phases.
During this phase, we introduce further degradations, such as adding stains and scribbles, casting shadows, and simulating document folding.
Examples of the resulting images are shown in Fig. \ref{fig:example_noise}.

\subsection{Final Dataset Construction}

We constructed the training set by combining the clean synthetic document images (prior to noise application), the scan-simulated images, and the Augraphy-processed images in proportions of 40\%, 30\%, and 30\%, respectively.
After manually filtering out images with synthesis errors or potential copyright concerns, the final dataset comprises a total of 17,970 images.

\section{Experiments}

\subsection{Experimental Setup}
\label{ssec:setup}

\subsubsection{Fine-Tuning Models on Our Dataset}

We fine-tune several open-source LVLMs using our dataset.
Specifically, we employ five models capable of reading Japanese text within images: Qwen2.5-VL-7B-Instruct\footnote{https://huggingface.co/Qwen/Qwen2.5-VL-7B-Instruct} \cite{bai2025qwen25vltechnicalreport}, Qwen3-VL-8B-Instruct\footnote{https://huggingface.co/Qwen/Qwen3-VL-8B-Instruct} \cite{bai2025qwen3vltechnicalreport}, InternVL3-8B\footnote{https://huggingface.co/OpenGVLab/InternVL3-8B-hf} \cite{zhu2025internvl3exploringadvancedtraining}, InternVL3.5-8B\footnote{https://huggingface.co/OpenGVLab/InternVL3\_5-8B-HF} \cite{wang2025internvl35advancingopensourcemultimodal}, and Gemma 3 12B IT\footnote{https://huggingface.co/google/gemma-3-12b-it} \cite{gemmateam2025gemma3technicalreport}.
In this study, we perform full fine-tuning by updating the parameters of all modules in these models.
During fine-tuning, we utilized a prompt instructing the model to output all text within the image as input.
We trained the model to generate the entirety of the text following the standard Japanese reading order.
The batch size is set to 32, utilizing the AdamW~\cite{loshchilov2018decoupled} optimizer with a learning rate of 2e-05.

\subsubsection{Baselines}

In this study, we compare our dataset against two baseline synthetic document image datasets.
The first baseline is the JSSODa dataset \cite{sasagawa2025evaluatingmultimodallargelanguage}.
Images in this dataset consist of simple synthetic documents featuring black text on a white background.
They encompass both horizontal and vertical writing directions, with layouts ranging from one to four columns.
Notably, no external images or figures are inserted within these document images.
Since Synth-JDoc is constructed from the text of JSSODa, the scale of both datasets is comparable.

The second baseline employs a high-performance text-to-image model to synthesize document images.
In the prompts provided to the model, we specify the same text as Synth-JDoc to be rendered within the document (including the title and figure captions generated in Section~\ref{ssec:element_preparation}), the writing direction, and the number of columns.
Furthermore, if figure captions are present in the text, we instruct the model to generate the corresponding figures.
We utilize Nano Banana Pro\footnote{We used \texttt{gemini-3-pro-image-preview}.} \cite{google_nanobananapro_2025} as the text-to-image model.
We excluded a portion of the data for which the image generation requests were rejected.
The scale of this dataset is comparable to that of Synth-JDoc.

By keeping the text identical between each baseline and Synth-JDoc, we can compare their performance while minimizing the effect of text distribution.
We fine-tuned the aforementioned five models using each of these datasets.
We kept the training configurations identical to those used when training on our dataset, and the volume of training data was kept consistent across all fine-tuning sessions.

\subsubsection{Evaluation Settings}

We evaluate the baseline models and the models fine-tuned on our dataset using the VJRODa dataset \cite{sasagawa2025evaluatingmultimodallargelanguage}.
The VJRODa dataset is a document OCR dataset for vertically written Japanese text, constructed from real-world document images.
This dataset consists of pairs of document images and their corresponding texts, comprising a total of 100 images.
For all models, the texts were generated via greedy decoding, utilizing prompts identical to those used during fine-tuning.

In this study, we follow the evaluation protocol of previous work \cite{sasagawa2025evaluatingmultimodallargelanguage}.
We employ Character Error Rate (CER) and BLEU as our evaluation metrics.
CER is calculated by dividing the edit distance between the model's output and the ground truth text by the number of characters in the ground truth, multiplied by 100.
For the BLEU calculation, we utilize SacreBLEU \cite{post-2018-call}, tokenizing both the model's output and ground truth texts at the character level.
Prior to calculating these scores, we apply Unicode NFKC normalization to the texts and remove all whitespace characters.

Furthermore, we evaluate the model outputs under two distinct configurations: the Raw Output setting and the Remove Repetition setting.
In the Raw Output setting, we calculate the scores using the model's output exactly as generated.
In the Remove Repetition setting, we calculate the scores after removing any trailing repetitive strings from the output.
The motivation for employing these two settings stems from the tendency of LVLMs to occasionally generate the same string repeatedly.
When this phenomenon occurs, fine-tuning might lead to an apparent score improvement in the Raw Output setting simply by mitigating this repetition, making it difficult to accurately evaluate the model's true character recognition performance.
Therefore, we also report the scores under the Remove Repetition setting.
Since both the ability to avoid repetitive generation and the core character recognition capability are crucial for LVLMs, it is vital to examine the results from both evaluation settings.

\subsection{Results}

\begin{table*}[t]
\centering
\footnotesize
\caption{
    The result on VJRODa.
    ``(+JSSODa)'', ``(+Nano Banana Pro)'', and ``(+\textbf{Ours})'' denote models fine-tuned on the JSSODa dataset, the document image dataset synthesized using Nano Banana Pro, and our constructed dataset (Synth-JDoc), respectively.
    Scores in rows marked with $^{*}$ are cited from prior work~\cite{sasagawa2025evaluatingmultimodallargelanguage}.
}
\label{table:eval_vjroda}
\begin{tabular}{lcccc}
\toprule
 & \multicolumn{2}{c}{Raw Output} & \multicolumn{2}{c}{Remove Repetition} \\
 \cmidrule(lr){2-3} \cmidrule(lr){4-5}
\textbf{Models} & CER($\downarrow$) & BLEU($\uparrow$) & CER($\downarrow$) & BLEU($\uparrow$) \\
\midrule
Qwen2.5-VL-7B-Instruct$^{*}$  & 154  & 20.1 & 88.5 & 22.0  \\ 
\quad (+JSSODa)$^{*}$         & 65.1 & 51.5 & 40.5 & 61.1 \\ 
\quad (+Nano Banana Pro)      & 161 & 20.9 & 135 & 24.4 \\ 
\quad (+\textbf{Ours})        & \textbf{34.5} & \textbf{66.8} & \textbf{32.0} & \textbf{69.5} \\ 
\midrule
Qwen3-VL-8B-Instruct          & 116 & 32.6 & 45.6 & 52.5 \\ 
\quad (+JSSODa)               & 130 & 29.9 & 65.5 & 49.4 \\ 
\quad (+Nano Banana Pro)      & 177 & 16.1 & 138 & 17.3 \\ 
\quad (+\textbf{Ours})        & \textbf{43.9} & \textbf{57.4} & \textbf{25.0} & \textbf{70.8} \\ 
\midrule
InternVL3-8B-hf$^{*}$         & 121  & 26.0 & 66.5 & 40.8 \\ 
\quad (+JSSODa)$^{*}$         & 251 & 26.1 & 73.5 & 54.9 \\
\quad (+Nano Banana Pro)      & 173 & 15.3 & 140 & 19.3 \\ 
\quad (+\textbf{Ours})        & \textbf{70.9} & \textbf{47.8} & \textbf{38.9} & \textbf{68.1} \\
\midrule
InternVL3.5-VL-8B-hf          & 121 & 29.2 & 56.1 & 41.0 \\ 
\quad (+JSSODa)               & 57.9 & 62.3 & 37.2 & 71.9 \\ 
\quad (+Nano Banana Pro)      & 173 & 15.6 & 117 & 18.8 \\ 
\quad (+\textbf{Ours})        & \textbf{36.8} & \textbf{66.7} & \textbf{25.9} & \textbf{78.3} \\ 
\midrule
Gemma 3 12B IT$^{*}$          & 125 & 17.5 & 67.9 & 23.3 \\ 
\quad (+JSSODa)$^{*}$         & \textbf{77.6} & \textbf{27.9} & \textbf{67.4} & \textbf{27.2} \\
\quad (+Nano Banana Pro)      & 196 & 8.51 & 145 & 9.26 \\ 
\quad (+\textbf{Ours})        & 128 & 18.7 & 96.3 & 26.5 \\ 
\bottomrule
\end{tabular}
\end{table*}

Table~\ref{table:eval_vjroda} presents the evaluation results.
Under both the Raw Output and Remove Repetition settings, the models fine-tuned on our dataset, excluding Gemma 3, achieved the best scores in both CER and BLEU.
These findings demonstrate that training models on our dataset most effectively enhances character recognition capabilities for real-world documents featuring complex layouts, including vertically written Japanese text.
For Qwen3-VL, InternVL3, and InternVL3.5, we observed a significant gap in scores between the Raw Output and Remove Repetition settings.
We attribute this discrepancy to the fact that the repetitive output behavior persisted even after fine-tuning on our dataset.

When trained on the dataset generated by Nano Banana Pro, all models exhibited degraded scores compared to their original, pre-trained versions (with the exception of the BLEU score for Qwen2.5-VL).
We believe this degradation is caused by the document images synthesized by Nano Banana Pro, which frequently contained distorted characters and failed to successfully generate multi-column layouts with vertically written text.

Training on Synth-JDoc did not improve the performance of Gemma 3.
We attribute this lack of improvement to how Gemma 3 handles input images.
Qwen2.5-VL and Qwen3-VL adopt Native Dynamic Resolution, while InternVL3 and InternVL3.5 employ a Dynamic Tiling Strategy.
These approaches allow them to process images with various aspect ratios and resolutions.
In contrast, Gemma 3 resizes input images to a fixed 1:1 resolution.
As shown in Figs. \ref{fig:example_synth_jdoc} and \ref{fig:example_noise}, the document images in Synth-JDoc encompass a wide range of resolutions.
Therefore, we hypothesize that Gemma 3 failed to learn effectively due to this fixed 1:1 resizing.
This suggests that enhancing OCR performance requires improvements not only in dataset quality but also in the model architectures.

\subsection{Analysis}

\subsubsection{Case Study}

\begin{figure}[t]
  \centering
  \begin{minipage}[c]{0.4\linewidth}
    \includegraphics[width=\linewidth]{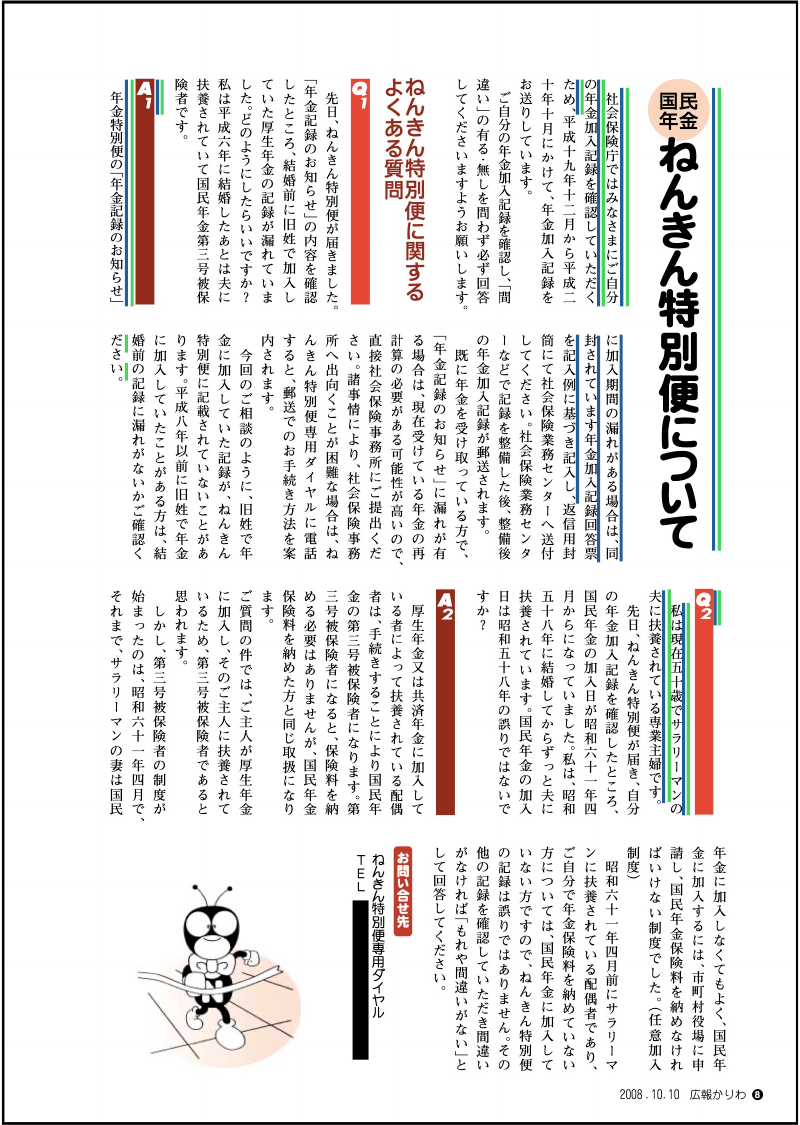}
  \end{minipage}
  \hfill
  \begin{minipage}[c]{0.59\linewidth}
      \footnotesize
      \begin{tabular}{@{} p{\linewidth} @{}}
        \toprule
        \textbf{Qwen3-VL-8B-Instruct}, CER: 47.4 (\textcolor{green}{green line}) \\
        \midrule
        国民年金\textbackslash nねんきん特別便について\textbackslash n\textbackslash n社会保険庁ではみなさまにご自分の年金加入記録を確認していただくため、 ($\cdots$) A1\textbackslash n年金特別便の「年金記録のお知らせ」\textbackslash nだい。\textbackslash n\textbackslash nQ2\textbackslash n私は現在五十歳でサラリーマンの夫に扶養されている専業主婦です。 ($\cdots$) \\
        \midrule
        \quad \textbf{(+Synth-JDoc)}, CER: 3.84 (\textcolor{blue}{blue line}) \\
        \midrule
        国民年金ねんきん特別便について\textbackslash n\textbackslash n社会保険庁ではみなさまにご自分の年金加入記録を確認していただくため、 ($\cdots$) A1 年金特別便の「年金記録のお知らせ」に加入期間の漏れがある場合は、同封されています年金加入記録回答票を記入例に基づき記入し、 ($\cdots$) Q2 私は現在五十歳でサラリーマンの夫に扶養されている専業主婦です。 ($\cdots$)\\
        \bottomrule
      \end{tabular}
  \end{minipage} \\
  \caption{
  Case study of output results between the original Qwen3-VL-8B-Instruct and the model fine-tuned on the Synth-JDoc dataset, using an image from the VJRODa dataset.
  The \textcolor{green}{green line} represents the output text of Qwen3-VL-8B-Instruct, while the \textcolor{blue}{blue line} represents that of the fine-tuned model.
  While the original model skips the text in the second column from the top, the fine-tuned model does not skip it.
  (Image is from https://warp.ndl.go.jp/info:ndljp/pid/11712522/www.vill.kariwa.niigata.jp/open/ info/000000001\_0000000609.pdf, page 8, a portion of the image was redacted.)
  }
  \label{fig:case_study}
\end{figure}

Figure \ref{fig:case_study} presents the output results of Qwen3-VL-8B-Instruct and its fine-tuned version on the Synth-JDoc dataset, evaluated on an image from the VJRODa dataset.
The original Qwen3-VL-8B-Instruct skips the text in the second column from the top, jumping directly from ``A1'' in the first section to ``Q2'' in the third.
In contrast, the model trained on Synth-JDoc does not skip this text, resulting in an improved CER score.
Furthermore, the models fine-tuned on JSSODa and the dataset synthesized by Nano Banana Pro exhibited repetitive string outputs in addition to character recognition errors.

\subsubsection{Japanese Document Images Generated by Nano Banana Pro}

\begin{figure}[t]
  \begin{minipage}[c]{\columnwidth}
    \centering
    \includegraphics[keepaspectratio, width=0.8\columnwidth]{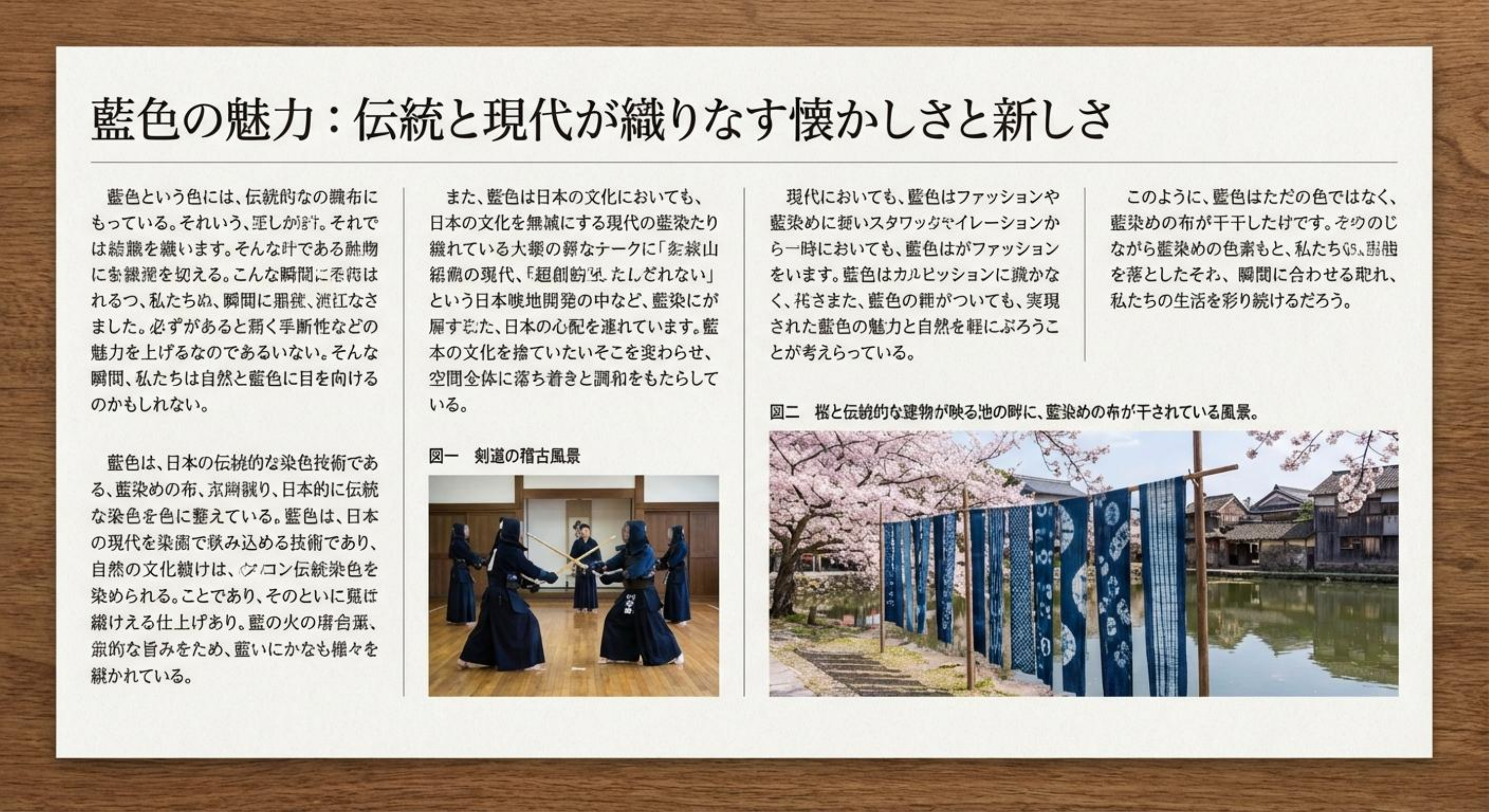}
  \end{minipage}\\
  \begin{minipage}[c]{\columnwidth}
    \centering
    \includegraphics[keepaspectratio, width=0.8\columnwidth]{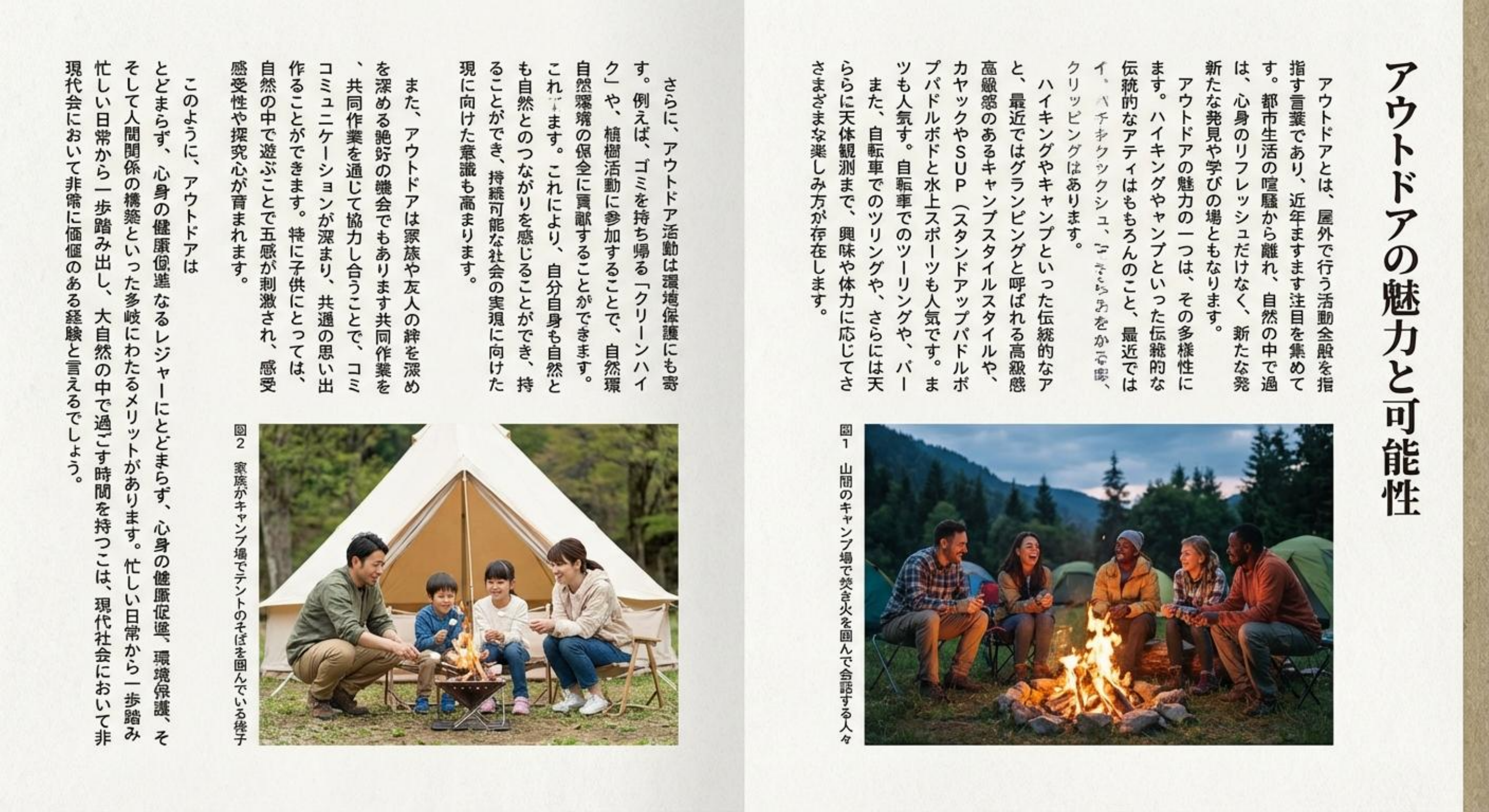}
  \end{minipage}\\
  \caption{
  Document images generated by Nano Banana Pro.
  The top and bottom images were generated with prompts specifying horizontal 3-column and vertical 2-column layouts, respectively.
  }
  \label{fig:example_nanobanana}
\end{figure}

Figure \ref{fig:example_nanobanana} shows examples of document images generated by Nano Banana Pro.
The top image was generated using instructions to produce a horizontally written, 3-column layout, whereas the bottom image was generated with instructions for a vertically written, 2-column layout.
An advantage of the images generated by Nano Banana Pro is their realistic appearance; for instance, they feature backgrounds like wooden desks and include natural details such as page folds.
However, the model has notable drawbacks, including distorted characters and text that differs from the content specified in the prompt.
Furthermore, it lacks the capability to generate multi-column documents with vertically written text.
As observed in the bottom image of Fig. \ref{fig:example_nanobanana}, although we instructed the model to generate a 2-column vertically written layout, it instead produced a document consisting of two separate vertically written pages.
We believe these limitations are the underlying causes for the lack of improvement in model performance.

\subsection{Ablation Study}
\begin{table*}[t]
\centering
\footnotesize
\caption{
    Ablation study on the effects of embedded images and captions in Synth-JDoc.
    ``w/o image/caption/image + caption'' indicates the scores of models trained on datasets where the respective elements are masked.
}
\label{table:ablation}
\begin{tabular}{lcccc}
\toprule
 & \multicolumn{2}{c}{Raw Output} & \multicolumn{2}{c}{Remove Repetition} \\
 \cmidrule(lr){2-3} \cmidrule(lr){4-5}
\textbf{Models} & CER($\downarrow$) & BLEU($\uparrow$) & CER($\downarrow$) & BLEU($\uparrow$) \\
\midrule
Qwen3-VL-8B-Instruct (+Synth-JDoc)   & \textbf{43.9} & \textbf{57.4} & \textbf{25.0} & \textbf{70.8}  \\
\quad (w/o image)                    & 60.5 & 48.1 & 36.8 & 67.4 \\ 
\quad (w/o caption)                  & 120 & 30.2 & 51.2 & 54.5 \\ 
\quad (w/o image + caption)          & 55.7 & 48.9 & 39.3 & 64.8 \\ 
\midrule
InternVL3.5-VL-8B-hf (+Synth-JDoc)   & \textbf{36.8} & \textbf{66.7} & \textbf{25.9} & \textbf{78.3} \\ 
\quad (w/o image)                    & 45.5 & 58.8 & 38.6 & 65.2 \\ 
\quad (w/o caption)                  & 54.3 & 57.9 & 39.9 & 72.7 \\ 
\quad (w/o image + caption)          & 59.3 & 57.6 & 47.1 & 66.4 \\ 
\bottomrule
\end{tabular}
\end{table*}

We conducted an ablation study to investigate the impact of embedded images and their corresponding captions on model performance.
First, prior to adding noise to the synthetic document images, we mask either the embedded images, their captions, or both by filling them with the background color.
Following this masking process, we apply the same noise used in the original Synth-JDoc.
Using the respective masked datasets, we train Qwen3-VL-8B and InternVL3.5-VL-8B, and compare their performance with that of the models trained on the original Synth-JDoc.

The evaluation results are presented in Table~\ref{table:ablation}.
The top row shows the results of the models trained on Synth-JDoc, while the subsequent three rows display the results of the models trained on datasets where either the embedded images, the captions, or both were masked, respectively.
For both Qwen3-VL-8B and InternVL3.5-VL-8B, the models trained on Synth-JDoc exhibited better performance than those trained on datasets with masked images or captions.
We attribute this to the fact that synthetic document images containing both embedded images and captions can represent more visually diverse and realistic documents.

\section{Limitations}
\subsection{Layout Diversity}

In this study, we synthesized document images with vertically and horizontally written multi-column layouts by creating HTML/CSS templates.
However, in reality, there are documents such as newspapers that feature layouts with more complex reading orders than those in the Synth-JDoc images.
Constructing datasets and models capable of handling such complex documents remains a subject for future work.

\subsection{Decoding Method}

During testing, we generated text using greedy decoding; however, we observed a phenomenon where the models repeatedly output the same strings.
As discussed in Section~\ref{ssec:setup}, the occurrence of this phenomenon can obscure the evaluation of the model's fundamental OCR capabilities.
Since adjusting decoding hyperparameters may suppress this repetitive behavior, exploring such techniques is a task for future work.

\subsection{Diverse Test Set}

In the Synth-JDoc construction pipeline, we applied noise to the synthesized document images using two distinct methods.
However, the VJRODa dataset used for our evaluation was constructed from PDFs published by government agencies, consisting of relatively clean images with minimal noise.
Consequently, we could not adequately evaluate the effectiveness of the noise application introduced in this study.
Constructing and evaluating on a noisy test dataset is a direction for future work.

\subsection{Removing NSFW Content}

The dataset construction pipeline in this study does not include filtering for Not Safe For Work (NSFW) content.
There is a possibility that NSFW content may be introduced during the stages of text preparation, prompt and image generation, and image captioning.
Implementing NSFW filtering for both text and images at each of these stages is crucial for constructing clean training datasets, and thus remains an important subject for future work.

\section{Conclusion}

In this study, we proposed Synth-JDoc, a synthetic Japanese document image dataset.
We prepared various elements, including text, title, images, and image captions, and utilized them to synthesize document images via HTML and CSS.
To construct more robust models, we applied noise to the generated document images.
Our results demonstrated that models trained on Synth-JDoc more effectively improved OCR capabilities on vertically written Japanese document images compared to those trained on other baseline datasets.
For future work, we plan to explore synthesis methods for document images incorporating complex elements such as graphs and tables.

\begin{credits}

\subsubsection{\ackname} 
This work was supported by the ``Development Acceleration Use'' program of ABCI 3.0, which is provided by AIST and AIST Solutions~\cite{takano2024abci30evolutionleading}.
In this work, we used the ``mdx: a platform for building data-empowered society''~\cite{9927975mdx}.

\subsubsection{\discintname}

The authors have no competing interests to declare that are relevant to the content of this article.
\end{credits}

\bibliographystyle{splncs04}
\bibliography{mybibliography}

\end{document}